\pdfoutput=1

\documentclass[11pt]{article}

\usepackage[final]{acl}

\usepackage{times}
\usepackage{latexsym}
\usepackage{algorithm} 
\usepackage{algpseudocode} 
\usepackage{amsmath}
\usepackage{bm}
\usepackage[T1]{fontenc}

\usepackage[utf8]{inputenc}

\usepackage{microtype}

\usepackage{inconsolata}

\usepackage{graphicx}
\usepackage{amsmath}
\usepackage{amssymb}
\usepackage{booktabs}
\usepackage{multirow}

\title{DAST: Context-Aware Compression in LLMs via Dynamic Allocation of Soft Tokens}

\author{
Shaoshen Chen$^{1}$,
~Yangning Li$^{1,2}$,
~Zishan Xu$^{1}$, 
~Yinghui Li$^{1}$,\\ 
~\textbf{Xin Su}$^{3}$,
~\textbf{Zifei Shan}$^{3}$,
~\textbf{Hai-Tao Zheng}$^{1,2}$\thanks{Corresponding author.},
 \\$^{1}$Shenzhen International Graduate School, Tsinghua University \\ 
          $^{2}$Peng Cheng Laboratory,$^{3}$WeChat,Tencent
         \\
        \tt{css24@mails.tsinghua.edu.cn}
}

\begin{document}
\maketitle
\begin{abstract}

Large Language Models (LLMs) face computational inefficiencies and redundant processing when handling long context inputs, prompting a focus on compression techniques. While existing semantic vector-based compression methods achieve promising performance, these methods fail to account for the intrinsic information density variations between context chunks, instead allocating soft tokens uniformly across context chunks. This uniform distribution inevitably diminishes allocation to information-critical regions. To address this, we propose Dynamic Allocation of Soft Tokens (DAST), a simple yet effective method that leverages the LLM’s intrinsic understanding of contextual relevance to guide compression.      DAST combines perplexity-based local information with attention-driven global information to dynamically allocate soft tokens to the informative-rich chunks, enabling effective, context-aware compression.  Experimental results across multiple benchmarks demonstrate that DAST surpasses state-of-the-art methods.\footnote[1]{We will release our code upon the acceptance of our paper.} 
\end{abstract}

\section{Introduction}
Large Language Models (LLMs) \cite{achiam2023gpt,bai2023qwen,touvron2023llama} have demonstrated remarkable performance on long context tasks, excelling capturing complex dependencies and generating coherent responses over extended contexts. Nevertheless, processing long contexts incurs high computational cost, making the development of efficient \textbf{context compression} methods that preserve semantic integrity while reducing input length crucial.

Early approaches to context compression primarily relied on context pruning or summarization \cite{jiang2023llmlingua,LLMlingua-2}, which reduced input length through content removal or rephrasing. However, these methods often compromise semantic integrity through direct modification of the input sequence. Recent semantic vector-based methods \cite{ge2024incontext,zhang2025long} address this limitation by replacing the original context of length $n$ with $m$ compressed soft tokens ($m\ll n$), preserving essential information in a more compact representation. Although effective, these methods typically append soft tokens at the context terminus or distribute them uniformly, overlooking uneven information density across context chunks. This uniform distribution prevents optimal allocation of compression capacity to information-rich regions.

Notably, text-pruning-based approaches like LongLLMLingua \cite{jiang-etal-2024-longllmlingua} attempt dynamic pruning using external models to estimate tokens importance. However, this external guidance fails to capture the LLM's intrinsic understanding of information relevance, creating incompatibility with vector-based methods.

This raises a key research question: \textbf{How can we dynamically allocate compression tokens based on the LLM’s inherent understanding of contextual information density?}

To address this, we propose \textbf{D}ynamic \textbf{A}llocation of \textbf{S}oft \textbf{T}okens (\textbf{DAST}), a simple yet effective approach to soft tokens compression that fully leverages the LLM’s internal capabilities without requiring external models. DAST utilizes perplexity to assess local importance and attention mechanisms to capture global relevance, dynamically allocating soft tokens based on intrinsic information density. This enables more efficient and context-aware compression, improving both compression quality and model performance compared to prior methods.

\begin{figure*}[h]
\centering
\includegraphics[width=\linewidth]{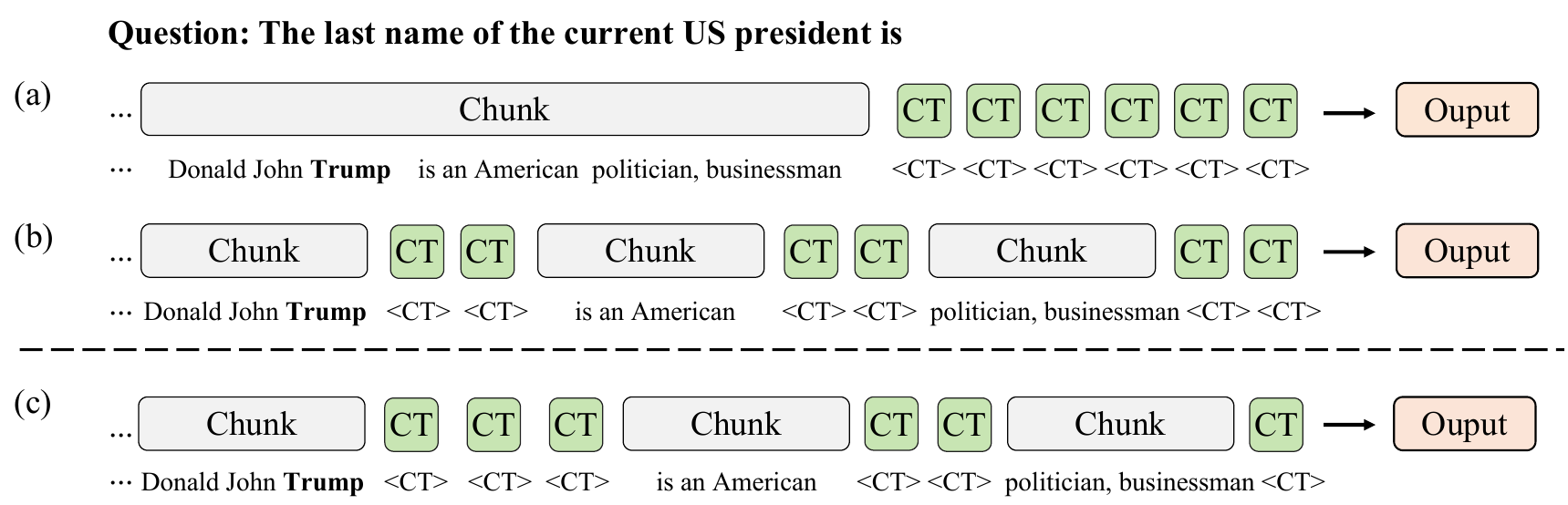}
\caption{(a) and (b): the previous \textbf{fixed} allocation methods include \textbf{Single-chunk} and \textbf{Multi-chunks compression}; (c): our \textbf{dynamic} allocation method. Notably, our dynamic method allocated more soft tokens to the key information in the answer (highlighted in \textbf{bold}) while reducing soft tokens for less information. $\text{<CT>}$ is compress soft token.
}
\label{tab:main_fig}
\end{figure*}

\section{Method}
\subsection{Compression Background}

Traditional methods for compressing long context sequences typically employ chunk-based decomposition. Given an input sequence \( X = \{X^{\text{que}}, X^{\text{doc}}\} \), where \( X^{\text{que}} \) denotes a query or instruction and \( X^{\text{doc}} \) represents a lengthy document, the sequence is segmented into \( N \) contiguous chunks of fixed length \( |\boldsymbol{X}_i| = L \). During compression, each chunk of length \( L \) is condensed into a fixed number \( m \) of soft tokens, where \( m \ll L \).

Existing compression strategies, as illustrated in ~\autoref{tab:main_fig}(a) and (b), can be broadly categorized into two paradigms. The first, termed \textbf{\textit{Single-chunk Compression}}, processes the entire sequence as a single chunk and appends all fixed \( m \) soft tokens after the full sequence. To enhance granularity, methods such as AutoCompress \cite{chevalier2023adapting} and Beacon \cite{zhang2025long} introduced \textbf{\textit{Multi-chunks Compression}}, which assigns a compression constraint (divisible by the chunk length \( L \)) stochastically during training and evenly distributes a fixed \( m \) soft tokens across all chunks during inference. 

However, a major limitation of these methods is their \textbf{fixed tokens allocation scheme}, which implicitly assumes uniform information density across the entire context. This assumption introduces the risk that regions with high information density receive fewer soft tokens, while regions with low information density are allocated more soft tokens. To address this issue, we propose a \textit{Dynamic Allocation of Soft Tokens} method, which adaptively assigns a \textbf{dynamic number of soft tokens} \( d_i \) to each chunk \( \boldsymbol{X}_i \), where \( d_i \) is determined by localized and global information density, as shown in ~\autoref{tab:main_fig}(c).



\subsection{Overall Framework}

Our method dynamically determines the number of soft tokens assigned to each chunk, as described in the next section. Given the \( i \)-th chunk, the compressed representation is constructed as:
\begin{equation}
\begin{aligned}
    \boldsymbol{C}_i = \left\{
    \boldsymbol{\langle ct \rangle}_1, \dots, \boldsymbol{\langle ct \rangle}_{i-1}, \, \boldsymbol{X}_i, \, \boldsymbol{\langle ct \rangle}_i
    \right\},
\end{aligned}
\end{equation}
where \( \boldsymbol{\langle ct \rangle}_j \in \mathbb{R}^{d_j} \) represents the compressed soft tokens of the \( j \)-th chunk (\( 1 \leq j < i \)). The compressed tokens \( \boldsymbol{\langle ct \rangle}_i \in \mathbb{R}^{d_i} \) capture essential information from the current chunk through contextual interactions, which are facilitated by the cross-attention mechanism:
\begin{equation}
    \text{CrossAttn.}(\boldsymbol{C}_i ; \text{Mask}).
\end{equation}

\subsection{Dynamic Allocation}
In this section, To effectively allocate soft tokens, we consider both \textbf{local importance} (i.e., within each chunk) and \textbf{global importance} (i.e., across the entire sequence). Specifically, we employ \textbf{Perplexity (PPL)} to estimate local importance and \textbf{Attention (Attn)} to capture global importance. These two metrics are then combined to determine the number of soft tokens assigned to each chunk.

\textbf{PPL}: Perplexity is a widely used metric for evaluating contextual informativeness. A lower perplexity signifies greater relevance of the current context information \cite{jiang2023llmlingua}. Since each chunk is visible within its own local context during compression, we compute the perplexity for each chunk separately. The resulting perplexity scores thus embody the local importance information of each respective chunk. The PPL of $i$-th chunk as follows:
\begin{equation}
\begin{aligned}
\text { ${P}_{i}$ = $- \sum_{l=1}^{L} q(x_l) \log p(x_l | x_{<l})$},
\end{aligned}
\end{equation}
where \( q(x_{l}) \) represents the probability distribution of the ground truth.

\textbf{Attn}: Once the sequence has been compressed, the importance of each chunk’s compressed representation can be inferred from attention weights.
Intuitively, chunks with more crucial information have higher attention weights, reflecting their contribution to global understanding. 
Hence, we utilized global attention to measure the global information, which has been demonstrated to be an effective method \cite{globalAttn}. Through attention weights, we can ascertain the global proportion of each token. The formula is :
\begin{equation}
\begin{aligned}
\text { $A_{i}$ = 
$\sum_{j=i}^{(i+1)m} (q  k^\top)_{j}$,
 }
\end{aligned}
\end{equation}
where \textbf{$q$} represents the last token vector and \textbf{$k$} represents all the compressed tokens vectors.
Thus, we can obtain the scores of i-th chunk that focus on both global and local information: 
\begin{equation}
\begin{aligned}
\text { $
\text{$S_{i}$} = \text{$A_{i}$} \cdot \alpha - \frac{\text{$P_{i}$}}{\sum_{k=1}^{N} (\text{$P_{k}$})} \cdot (1 - \alpha)$,
}
\end{aligned}
\end{equation}
where $\alpha$ is a parameter that balances the importance of global \(A_{i}\) and local \(P_{i}\) information. Then Softmax is used for normalization. Noted that since \( A_{i} \) and \( P_{i} \) are derived from different distributions,  \( P_{i} \) is scaled by the number of all chunk \( N \). Given the total number of soft tokens of context is \( M \), we can calculate the actual number of soft tokens in the \( i \)-th chunk $d_{i}$ = $M$ $\times$ $S_{i}$.

\textbf{Reallocation}: In order to make the tokens after dynamic allocation consistent with the training, we design a reallocation algorithm to make them divisible by $L$. The details of this algorithm are shown in Appendix~\ref{appendix:algorithm}. It is worth mentioning that the reallocation is optional, depending on the method mechanism used.

\section{Experiments}
The experiment's detail  are presented in Appendix~\ref{appendix:implementation}. Next, we want to answer two questions: (1) How effective is DAST? (2) How does the performance of DAST improve?
\subsection{Main Results}
To evaluate the dynamic distribution capability of DAST in a long context with inconsistent ground truth granularity distribution, we employ three benchmarks from LongBench \cite{bai-etal-2024-longbench}: Single-Document, Multi-Document, and Example tasks (Few-Shot). As demonstrated in ~\autoref{tab:main_longbench}, our approach demonstrates consistent superiority over baseline methods across all evaluated tasks. This improvement can be attributed to the model's adaptive capacity to discern textual saliency within redundant, extended textual inputs. Our method strategically allocates a higher proportion of soft tokens to semantically critical chunks, thereby enhancing computational attention to pivotal content. This differential allocation mechanism ultimately optimizes task-specific performance through context-aware resource distribution.

\begin{figure}[tb]
\centering
\includegraphics[width=\linewidth]{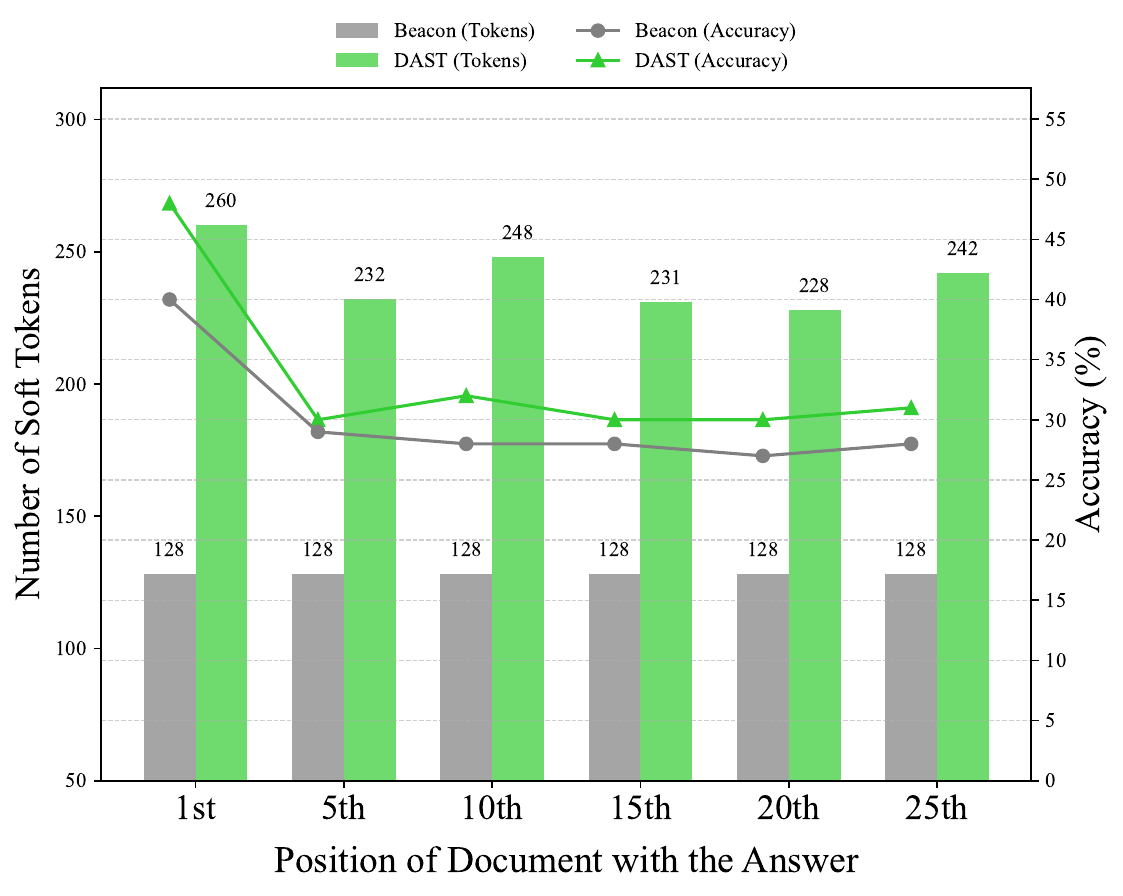}
\caption{Performance and Number of soft tokens v.s. Key Information Position.}
\label{NQ}
\end{figure}

Having established the effectiveness of our dynamic compression method, we further examine its performance-enhancing mechanisms through systematic experiments on the NaturalQuestions dataset~\cite{NQ}. This benchmark is particularly well-suited for analysis due to the presence of correct answers at varying contextual positions. As shown in ~\autoref{NQ}, our method consistently surpasses the uniform compression approach of Beacon across all positional configurations. Notably, in context chunks containing answer-relevant, our method adaptively allocates more tokens to these semantically critical regions, thereby improving performance.

\begin{table*}[h]
\centering
\begin{tabular}{lcccc|cccc}
\toprule
                & \multicolumn{8}{c}{\textbf{Document and Example Compression}}                                                \\ \cmidrule(lr){2-9}
\multirow{2}{*}{\textbf{Methods}}         & Single & Multi & Few  & \multirow{2}{*}{\textbf{AVG}} & Single & Multi & Few  & \multirow{2}{*}{\textbf{AVG}}   \\ 
&Doc & Doc & Shot  &  & Doc & Doc & Shot  & 
\\ \midrule
& \multicolumn{4}{c}{\textbf{LLama-2-7B}}   & \multicolumn{4}{c}{\textbf{Qwen-2-7B}}                                                                                                       \\ \midrule
Original Prompt
& 24.9           & 22.5               & 60.0             &  35.8   & 22.0         & 29.3               & 62.3             &  37.9                           \\
Zero-Shot       &   8.1             & 6.1    & 32.2 & 15.5          & 7.1           & 6.6               & 26.8             &  13.5                       \\ \midrule                                                
AutoComp.$^{\dag}$ \citep{chevalier2023adapting}                    & 12.9      & 16.4     & 23.8     &  17.7         &\multicolumn{4}{c}{-}                 \\
ICAE$^{\dag}$ \citep{ge2024incontext}                               & 19.5      & 19.2       & 24.8    & 21.2                &\multicolumn{4}{c}{-}                 \\
LongLingua.$^{\dag}$ \citep{jiang-etal-2024-longllmlingua}                     & 21.5      & 18.8       & 49.5     &     29.9      & 24.7      & 20.3       & 55.9     &     33.6                    \\
SnapKV$^{\dag}$ \citep{li2024snapkv}                            & 24.2      & 22.6       & 60.1     &   35.6         & 38.7      & 37.6            &   67.1   &47.8                \\
Beacon$^{\dag}$ \citep{zhang2025long}                      & 34.9      & 27.5       & 61.4     &  41.3                  & 40.5      & 40.3       & 68.4     &   49.7             \\
\textbf{DAST (Ours)}   &  \textbf{38.1}             &  \textbf{37.4}       &          \textbf{63.6}   & \textbf{46.4}  & \textbf{40.6} & \textbf{45.6} & \textbf{68.6} &  \textbf{51.6}                 \\ \bottomrule
\end{tabular}
\caption{Evaluation of various Document and Example Compression tasks (top performances marked in $\textbf{bold}$). $\dag$: the results cited from \citet{zhang2025long}.}
\label{tab:main_longbench}
\end{table*}


\subsection{Comparison of Different Constraints}
We compare our method to baselines on the Long-Term Memory MSC dataset \cite{MSC} to study compression intensity vs. memory retention. As presented in ~\autoref{ratios}, our approach consistently outperforms conventional methods across all compression levels, especially in resisting degradation at higher compression constraints.

\begin{table}[tb]
\resizebox{\linewidth}{!}{
\begin{tabular}{lcccc}
\toprule
\multirow{2}{*}{\textbf{Method}} & \multicolumn{4}{c}{\textbf{Compression Constraint}}                                                                                                                                                                                                                                                                                                                                                                 \\ \cmidrule(lr){2-5} 
                                 & \begin{tabular}[c]{c} $\sim$4 x\end{tabular}
                                 & \begin{tabular}[c]{c} $\sim$8 x\end{tabular}  
                                 & \begin{tabular}[c]{c} $\sim$16 x\end{tabular}
                                 & \begin{tabular}[c]{c} $\sim$24 x\end{tabular} \\  \midrule
AutoCom.                 &    28.8                                                                &     27.3$_{\downarrow5.2\%}$                                                                &                25.0$_{\downarrow13.2\%}$                                                                        &    24.0$_{\downarrow17.1\%      }$                                                                \\
LongLLM.                    &     22.4                                                               &   19.5$_{\downarrow13.0\%  }$                                                                 &      17.8$_{\downarrow20.5\%   }$                                                                                                                                       &   15.9$_{\downarrow29.0\%  }$                                                                     \\
ICAE                    &     18.1                                                               &    16.6$_{\downarrow8.3\% }$                                                                   &     15.3$_{\downarrow15.5\% }$                                                                                                                                          &   14.5$_{\downarrow19.9\%}$                                                                      \\
Beacon                    &  39.0                                                                  &  36.5$_{\downarrow6.4\%}$                                                                      &    33.6$_{\downarrow13.9\%}$                                                                                    & 32.3$_{\downarrow17.2\%}$                                                                         \\
\textbf{DAST}                          &        $\textbf{55.9}$                                                            &            $\textbf{55.5}_{\downarrow0.7\%}$                                                                                                                               &    $\textbf{52.6}_{\downarrow5.9\%}$                                                                   &         $\textbf{51.9}_{\downarrow7.2\%}$                                                                                                     \\ \bottomrule                                                                
\end{tabular}
}
\caption{Evaluation of Long-Term Memory on MSC.  $\downarrow$: percentages showing relative performance drop compared to the $\sim$4x compression baseline.}
\label{ratios}
\end{table}

\subsection{Ablation Study}

\begin{table}[t]
\centering
\begin{tabular}{lc}
\toprule
\textbf{Method}          & \textbf{Single-Doc} \\ \midrule
Random Allocation & 34.52  \\ \midrule
Uniform Allocation  &      34.90              \\ \midrule
\textbf{Dynamic Allocation (ours)}                    &      \textbf{38.14}         \\
w/o PPL       &       37.60              \\
w/o Attn      & 37.24                   \\
\bottomrule 
\end{tabular}
\caption{Ablation study of DAST.}
\label{ablation}
\end{table}
In this section, we analyze our method's performance and the impact of each module (see ~\autoref{ablation}). Random and uniform tokens allocations performed poorly, showing insufficient focus on critical context segments. Removing either the global attention (Attn) or local perplexity (PPL) module individually
caused performance drops, highlighting their importance in
prioritizing important chunks.

\begin{figure}[tbh]
\centering
\includegraphics[width=\linewidth]{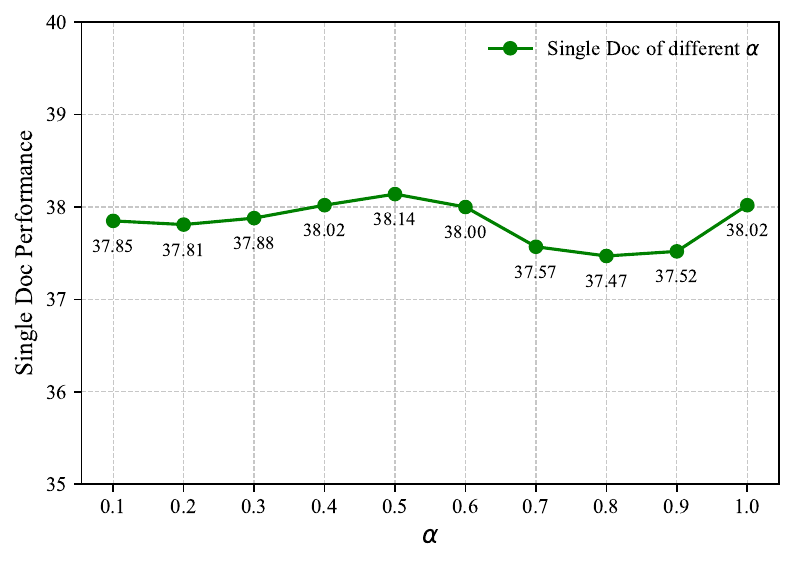}
\caption{Parameter Sensitivity Analysis of $\alpha$.}
\label{alpha}
\end{figure}

\label{sec:3.4}
\subsection{Parameter Sensitivity Analysis}
It is important to analyze the sensitivity of the parameters $\alpha$. 
As shown in ~\autoref{alpha}. The results indicate that the performance remains stable across different values of $\alpha$. Consequently, we selected a default value of $\alpha=0.5$ in main experiments, which simplifies the application of our method to other models, as it eliminates the need for specialized parameter tuning.

\section{Conclusion}
In this paper, we propose DAST, a simple yet effective  method that dynamically allocates soft tokens by leveraging the LLM’s intrinsic perception of information density. By integrating perplexity-based local information and attention-driven global relevance, DAST adaptively focuses compression capacity on high-information regions without relying on external models. 
Our experiments show that DAST outperforms prior methods in both compression quality and downstream task performance, underscoring the value of model-guided dynamic allocation. 

\section{Limitations}
Current model compression research remains primarily limited to the approximately 7B parameter scale due to computational resource constraints. While our study has demonstrated that our method outperforms other compression approaches, we have not been able to systematically investigate whether existing compression techniques, including our approach, can maintain their effectiveness when applied to larger architectures.  Given that the practical value of compression techniques becomes more pronounced with increasing model sizes, this represents a critical direction for future research.  Furthermore, it is also imperative to examine whether the compression process induces more severe : (1) hallucination phenomena and (2) catastrophic forgetting in compressed models, which constitutes another essential aspect requiring thorough investigation.


\bibliography{custom}

\appendix

\section{Reallocation Algorithm}
For the current soft tokens set $\bm{T}$ of all chunks given, the total soft tokens $\bm{S}$ can be assigned, and the optional compression rate set $\bm{R}$, our goal is to reassign $\bm{T}$ to get $\tilde{\bm{T}}$, the algorithm is as follows Algorithm~\autoref{new_algorithm}.
\label{appendix:algorithm}
\begin{algorithm} 
	\caption{Reallocation} 
	 \label{algorithm}       
	\begin{algorithmic}[1] 
	\Require $\bm{T}$, $\bm{S}$,$\bm{R}$ ,   
    \Ensure Reallocated $\tilde{\bm{T}}$,
    \State $\tilde{\bm{T}}$ $\gets$ $\emptyset$   
    \State ${\bm{T}}$ are allocated to each chunk based on the closest compression constraint ${\bm{R}}$ to get the $\tilde{\bm{T}}$.
    \Repeat
    \State Get the disposable ${\bm{M}}$ from ${\bm{S}}$ - ${\bm{sum(\tilde{T})}}$
    \State In the remaining tokens, double the tokens of the chunk that meets the conditions with the highest score.
    \State Update $\tilde{\bm{T}}$
    \State Remove the highest score temporarily,
 
   \Until {$\bm{M}$ = 0}
   \State\Return $\tilde{\bm{T}}$ 
   \end{algorithmic} 
\label{new_algorithm}
\end{algorithm} 

\section{Settings}
\label{appendix:implementation}
\subsection{Implementation}

To ensure strict methodological consistency in model comparisons, we employ the Llama-2-7B (chat) \cite{llama2chat} and Qwen-2-7B \cite{bai2023qwen} architectures. For training data selection, we adopt the same approach as Beacon\cite{zhang2025long}, utilizing 1B tokens sampled from Repajama \cite{redpajama} during pre-training, supplemented by LongAlpaca \cite{chen2024longlora}, BookSum \cite{kryscinski2022booksum}, and synthetic data generated by GPT-3.5 for fine-tuning (see Beacon \cite{zhang2025long} for detailed data curation protocols).
Our implementation uses a standard $\alpha$ value of 0.5, with sensitivity analyses for alternative parameter configurations provided in \hyperref[sec:3.4]{\textsection 3.4}. We use the HuggingFace framework \cite{huggingface} and all experiments were conducted on a computational cluster equipped with 8 $\times$ A800 GPUs (80GB).
\subsection{Baselines}
We compare our method to a baseline (represented by the Original Prompt) with the same constraints and an uncompressed baseline with no long context data (represented by Zero-Shot). In addition, We also compared with the current mainstream including text pruning or summarization long context compression method and semantic vector-based long context compression methods.
These include AutoCompressors\cite{chevalier2023adapting}, ICAE\cite{ge2024incontext}, LongLLMLingua\cite{jiang-etal-2024-longllmlingua}, SnapKV\cite{li2024snapkv}, and Beacon\cite{zhang2025long}.
\end{document}